\documentclass[11pt]{article}

\usepackage[preprint]{acl}

\usepackage{times}
\usepackage{latexsym}

\usepackage[T1]{fontenc}

\usepackage[utf8]{inputenc}

\usepackage{microtype}

\usepackage{inconsolata}

\usepackage{graphicx}
\usepackage{booktabs}
\usepackage{amsmath}
\usepackage{multirow}
\usepackage[table]{xcolor}
\usepackage{amssymb} 
\usepackage{algpseudocode}
\usepackage{fancyvrb}
\usepackage{algorithm}
\usepackage{algpseudocode}
\usepackage{fvextra}
\usepackage[most]{tcolorbox}
\tcbuselibrary{listings} 
\usepackage{booktabs} 
\usepackage{pifont}   
\newcommand{\xmark}{\ding{55}} 
\usepackage{tabularx}
\usepackage{array}

\lstdefinelanguage{json}{
    basicstyle=\small\ttfamily,
    columns=fullflexible,
    keepspaces=true,
    breaklines=true,
    frame=none,
    aboveskip=0pt,
    belowskip=0pt
}

\title{Mem2ActBench: A Benchmark for Evaluating Long-Term Memory Utilization in Task-Oriented Autonomous Agents}

\author{
Yiting Shen\textsuperscript{1},
Kun Li\textsuperscript{1}, 
Wei Zhou\textsuperscript{1},
Songlin Hu\textsuperscript{1,2}\\
\textsuperscript{1}Institute of Information Engineering, Chinese Academy of Sciences, Beijing, China\\
\textsuperscript{2}School of Cyberspace Security, University of Chinese Academy of Sciences, Beijing, China
}

\begin{document}
\maketitle

\begin{abstract}
Large Language Model (LLM)-based agents are increasingly deployed for complex, tool-based tasks where long-term memory is critical to driving actions. Existing benchmarks, however, primarily test a angent's ability to passively retrieve isolated facts in response to explicit questions. They fail to evaluate the more crucial capability of actively applying memory to execute tasks. To address this gap, we introduce \textsc{Mem2ActBench}, a benchmark for evaluating whether agents can proactively leverage long-term memory to execute tool-based actions by selecting appropriate tools and grounding their parameters. The benchmark simulates persistent assistant usage, where users mention the same topic across long, interrupted interactions and expect previously established preferences and task states to be implicitly applied.  We build the dataset with an automated pipeline that merges heterogeneous sources (ToolACE, BFCL, Oasst1), resolves conflicts via consistency modeling, and synthesizes 2,029 sessions with 12 user--assistant--tool turns on average. From these memory chains, a reverse-generation method produces 400 tool-use tasks, with human evaluation confirming 91.3\% are strongly memory-dependent. Experiments on seven memory frameworks show that current systems remain inadequate at actively utilizing memory for parameter grounding, highlighting the need for more effective approaches to evaluate and improve memory application in task execution. Code and data are available at \url{https://anonymous.4open.science/r/Mem2ActBench-29AC/}.



\end{abstract}
\section{Introduction}

\begin{figure}[t]
    \centering
    \includegraphics[width=\linewidth]{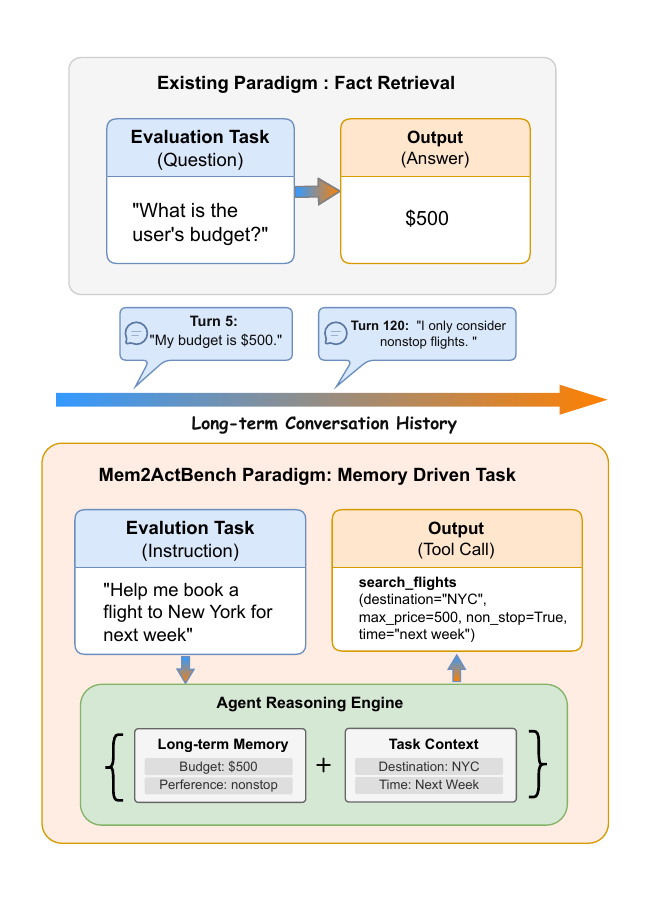}
     \caption{\textbf{Fact retrieval vs. memory-driven task execution.} Existing benchmarks focus on direct queries for a factual answer. In contrast, our benchmark requires the agent to combine past memories and  generate a grounded tool call.}
     
    \label{fig:paradigm}
\end{figure}

Large language model (LLM)-based agents are increasingly used as persistent assistants, interacting with users over extended periods. In these scenarios, users rarely restate all task constraints explicitly. Instead, preferences, requirements, and partial task states are gradually established across prior interactions, often interrupted by unrelated conversations, and are implicitly assumed to be remembered and applied in later requests. A realistic assistant is therefore expected not only to store long-term memory, but to actively retrieve and apply relevant past information to execute concrete actions, such as grounding missing arguments in tool invocations. Current memory benchmarks primarily  test an agent's ability to retrieve isolated information from memory based on explicit questions, such as MSC \cite{xu2022beyond} and Locomo \cite{maharana2024evaluating} (e.g., "What is the user's budget?"), but may under-test a more realistic challenge: given an underspecified instruction, the agent must infer what constraints to retrieve from long-term memory and ground them into an executable tool invocation, as shown in Figure~\ref{fig:paradigm}.

To bridge this gap, we introduce \textbf{\textsc{Mem2ActBench}}, a benchmark that evaluates whether agents can reconstruct executable tool arguments from dispersed long-term memory. Unlike prior benchmarks that test explicit memory retrieval, \textsc{Mem2ActBench} models scenarios where task is clear but critical execution constraints are distributed across long, interruption-heavy histories.Each instance is constructed so that the correct tool invocation is uniquely grounded in memory but cannot be inferred from the final query alone. We construct an automated pipeline to interleave task-oriented tool-use data with natural dialogue to construct long-term interaction histories that reflect realistic assistant usage. To transform these histories into usable long-term memory, we resolve conflicting states and organize extracted facts into a coherent memory evolution chain that captures how topics are updated over time, before applying reverse query generation with strict leakage control to ensure genuine memory dependence.

The main contributions of this work are:

\begin{itemize}
    \item We introduce a principled benchmark design for evaluating inference-driven long-term memory utilization in tool-augmented agents, targeting scenarios where task execution requires grounding underspecified requests using historical constraints.
    
    \item We construct and release \textsc{Mem2ActBench}, a benchmark comprising 400 memory-dependent tool-use tasks derived from 2,029 long-context dialogue sessions. Human verification confirms that 91.3\% of the tasks cannot be solved without access to long-term memory, ensuring the reliability of the evaluation.
    
    \item We conduct a comprehensive evaluation of seven representative memory frameworks and systematically analyze their failure modes, revealing persistent bottlenecks in memory retrieval and parameter grounding for tool-using tasks.
\end{itemize}

\section{Related Work}
\subsection{Agent Memory Architectures}

For autonomous agents in long-horizon, multi-stage tasks, memory has shifted from passive storage to an active module that supports planning and tool use. Existing approaches broadly fall into: (i) extending the context window to include long histories, which can suffer from the ``lost-in-the-middle'' effect and higher inference cost \cite{liu-etal-2024-lost}; (ii) external memory banks (e.g., vector stores) that retrieve stored interaction fragments or facts on demand, such as RET-LLM \cite{modarressi2024retllmgeneralreadwritememory} and MemoryBank \cite{zhong2024memorybank}; and (iii) explicit memory managers that organize and update memory structures, including Generative Agents \cite{park2023generative}, MemGPT \cite{packer2024memgptllmsoperatingsystems}, and A-Mem \cite{xu2025amemagenticmemoryllm}. Despite these advances, most evaluations still emphasize memory \emph{storage} and \emph{retrievability}, rather than whether agents can decide \emph{what} to retrieve and \emph{how} to apply it under tool and task constraints.

\begin{table*}
    \centering
    \resizebox{\linewidth}{!}{
        \begin{tabular}{lccccccc}
            \toprule
            \textbf{Dataset} & \textit{Session} & \textit{Turns} & \textit{Tokens} & \textit{QA pairs} & \textit{Reasoning} & \textit{Memory Evolution} & \textit{Tool use} \\
            \midrule
            MSC\cite{xu2022beyond} & 4 & 53& 564& / & Retrieval & \xmark & \xmark \\
            MemoryBank\cite{zhong2024memorybank} & 10 & 38& 3094& 7& Retrieval & \checkmark & \xmark \\
            Locomo\cite{maharana2024evaluating} & 27.2& 21.6 & 16618.1 & 199 & Retrieval & \checkmark & \xmark \\
            DialSim\cite{kim2025dialsimdialoguesimulatorevaluating} & 1313 & 1310 & 352k & 1056 & Retrieval & \checkmark & \xmark \\
            LongmemEval\cite{wu2024longmemeval} & 48 / 500 & 10.34 & 115k / 1.5M & 500 & Retrieval & \checkmark & \xmark \\
            \midrule
            \textbf{Mem2ActBench(Our work)} &  2029& 13& 3238& 400& Inference & \checkmark & \checkmark \\
            \bottomrule
        \end{tabular}
    }
    \caption{\textbf{Comparison of representative agent-memory benchmarks.}
    \textit{Session} is the number of discrete conversation segments per sample (temporal span);
    \textit{Turns} is the total dialogue turns (interaction length);
    \textit{Tokens} is the total token count aggregated over all sessions (text scale);
    \textit{QA pairs} is the number of question--answer instances used for evaluation;
    \textit{Reasoning} indicates whether tasks primarily test factual retrieval or inference;
    \textit{Memory Evolution} indicates whether memory states are dynamically updated during interaction;
    \textit{Tool use} indicates whether external tool invocation is supported in evaluation.}
    
    \label{tab:comparison}
\end{table*}

\subsection{Agent Memory Benchmarks}

Most memory benchmarks follow an explicit query-based paradigm. Early work targets short-dialogue consistency (e.g., Persona-Chat \cite{yamashita2023realpersonachat}), while MSC \cite{xu2022beyond} extends to cross-session long-term attribute retention. More recent benchmarks, such as LoCoMo \cite{maharana2024evaluating}, LongMemEval \cite{wu2024longmemeval}, and MemoryAgentBench \cite{hu2025evaluating}, increase time span and retrieval difficulty, but still largely instantiate ``Question $\rightarrow$ Retrieval $\rightarrow$ Answer'' with an explicitly provided query. This design under-tests realistic settings where retrieval intent must be inferred from underspecified task demands (e.g., missing tool arguments), rather than directly asked. Tool-oriented benchmarks similarly provide limited coverage of long-term memory usage: even approaches that incorporate memory into tool invocation, such as BFCL-v4 \cite{patil2025bfcl}, typically operate over short interaction horizons. 

As summarized in Table \ref{tab:comparison}, prior benchmarks emphasize explicit query memory matching and static memory usage. \textsc{Mem2ActBench} instead targets inference-driven long-term memory utilization, evaluating whether agents can infer task-critical constraints from evolving interaction histories and ground them into executable tool calls.

\section{Methodology}
\subsection{Overview}
To evaluate an agent's ability to proactively apply long-term memory for task execution, we introduce \textsc{Mem2ActBench}, constructed via a three-stage automated pipeline. First, we simulate realistic, interruption-heavy interactions by interleaving task-oriented data with conversational noise, creating fragmented contexts that necessitate long-term memory. Next, we synthesize these interactions into a logically coherent Fact Evolution Chain to serve as a ground-truth memory. Finally, we employ a reverse-generation paradigm, creating underspecified queries derived from ground-truth tool calls. This design ensures that successful task completion strictly requires reasoning over the historical memory chain, thereby directly evaluating the agent's ability to apply memory for inference-driven tasks rather than just retrieving facts.

\begin{figure*}[t]
    \centering
    \includegraphics[width=\textwidth]{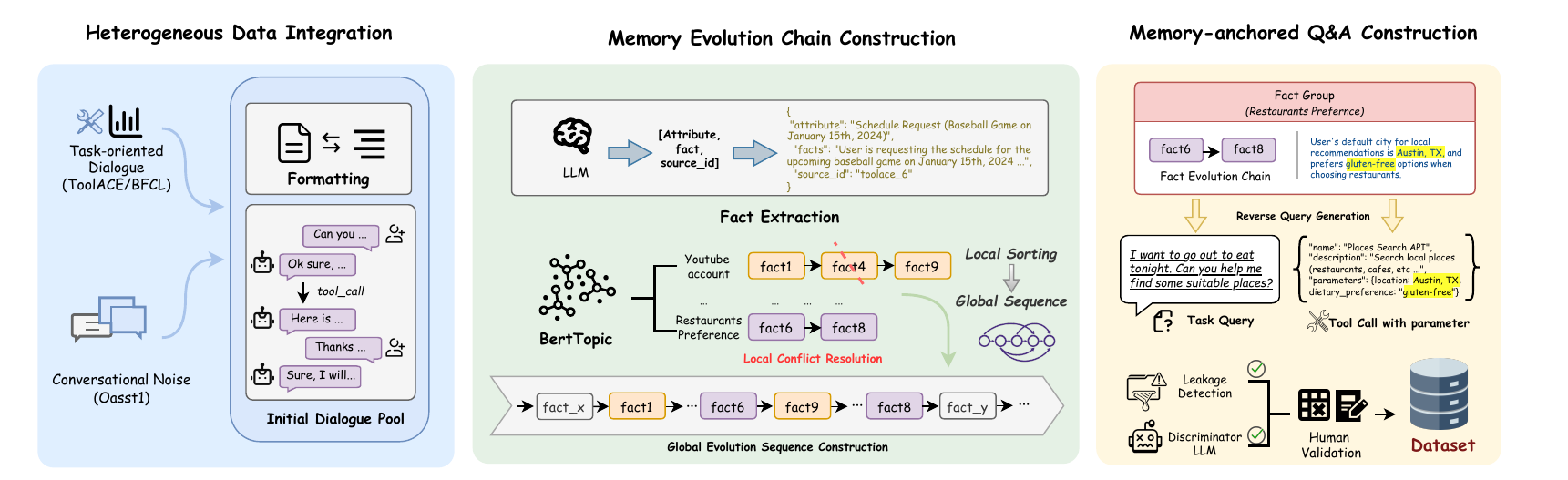}
    \caption{This diagram illustrates the \textsc{Mem2ActBench} framework, a benchmark used to evaluate the long-term memory capabilities of an agent. The framework first constructs a globally consistent and conflict-free "memory evolution chain" by integrating multi-source dialogue data. Then, based on this memory chain, it reverse-engineers question-answering tasks that require long-term memory to correctly select and use tools. Through this automated process, \textsc{Mem2ActBench} can effectively measure an agent's ability to proactively use its memory to complete tasks in complex, long dialogues.}
    \label{fig:framework}
\end{figure*}

\subsection{Heterogeneous Data Integration}

\paragraph{Task-oriented Dialogue.}
We construct the dataset by synthesizing multi-step tool-use trajectories from ToolACE and BFCL via LLM-based generation. For ToolACE\cite{ICLR2025_663865ea}, we process 8,000 samples, parsing raw interaction traces and employing LLM to reconstruct them into coherent, natural multi-turn dialogues. For BFCL\_v3\cite{patil2025bfcl}, we aggregate diverse subsets (including live, parallel, and multi-turn scenarios). Using the same synthesis approach, we transform static task queries and ground-truth tool calls into dynamic multi-round conversations, ensuring the dataset faithfully reflects realistic user-assistant interactions and precise tool execution flows.

\paragraph{Conversational Noise.}
We inject conversational noise from OASST1~\cite{kopf2023openassistant}, a tree-structured corpus with ranked assistant candidates. We keep only \texttt{rank=0} responses and reconstruct full threads by tracing selected leaves to the root.

We collect these dialogues and normalize them into a unified multi-turn format. After alignment, all processed interactions serve as the historical dialogue repository for subsequent memory construction and task generation. Details are provided in Appendix \ref{sec:data_process_details}. 

\subsection{Constructing the Fact Evolution Chain}


\subsubsection{Fact Extraction and Grouping}

We prompt an LLM to extract structured facts from each dialogue. Each fact is represented as a triple (\textbf{attribute}, \textbf{fact}, \textbf{source ID}), where \textbf{fact} is the atomic user statement and \textbf{source ID} uniquely identifies the originating dialogue. To prevent unrelated events from being merged under overly generic labels, we instruct the LLM to produce entity-bound attributes whenever applicable (e.g., “account modification (YouTube)”), which reduces spurious cross-entity comparisons. 

We then cluster the extracted \textbf{attributes} using \textbf{BERTopic}~\cite{grootendorst2022bertopic} with HDBSCAN as backend. For each cluster, we select one attribute as the canonical representative and map all attributes in the cluster to it, flagged as outliers. The resulting attribute clusters are used as \textbf{fact groups} for subsequent conflict detection and evolution analysis.

\subsubsection{Memory Evolution Chain Construction}


\paragraph{Local Conflict Resolution.}
For each fact group (sharing the same attribute), we use an LLM to produce a locally consistent evolution chain. Concretely, the LLM (i) orders facts by their temporal cues, (ii) preserves logically valid updates, such as refinement from coarse to specific (e.g., sports'' $\rightarrow$ basketball'') and valid multi-valued trajectories (e.g., residences over time), (iii) removes statements that are off-context or in strong logical conflict with other facts under predefined rules, and (iv) drops near-duplicates that provide no information gain. The remaining facts are then compressed into a clean local sequence, which serves as ordering constraints for global integration.

\paragraph{Global Evolution Sequence Construction.}
We merge the local sequences into one global evolution chain. This is achieved by first constructing a dependency graph where facts are nodes and temporal orderings are directed edges. Next, we apply a modified topological sorting method based on Kahn’s algorithm. To handle contradictions that manifest as cycles in the graph, we introduce a deterministic heuristic for conflict resolution. When a cycle is found, we identify the deadlocked nodes and remove the one with the highest out-degree. This removes the fact that forces the most downstream ordering constraints, which helps restore a valid order. The final outputs are the globally sorted sequence of facts and a list of any conflicting facts that were discarded.

\subsection{Memory-anchored Q\&A Construction}


\subsubsection{Target Tool Selection and Parameter Anchoring}
Given a memory evolution chain $\mathcal{S}$, we first construct a fully specified gold-standard tool invocation $C=(t^*, P)$ that is strictly grounded in memory.  The target tool $t^*$ is selected via hybrid retrieval (BM25 + BGE-M3) followed by LLM-based decision-making. For parameter construction, all values in $P$ must be either explicitly extracted from or logically inferred based on $\mathcal{S}$. To prevent spurious or hallucinated parameters, we enforce a memory-anchoring constraint: each parameter is validated through a combination of fuzzy matching and an LLM verifier, ensuring that its value can be traced back to the memory chain. 

\subsubsection{Reverse Implicit Query Generation and Filtering}

Starting from the grounded tool invocation $C$ and its supporting memory subset, we reverse-generate an underspecified user query $Q$. The generation process enforces three critical constraints:
(i) Parameter Omission: Key values present in $C$ must be omitted from $Q$ to prevent information leakage;
(ii) Reference Dependency: The query must rely on anaphoric expressions (e.g., "book \textit{that} flight", "use \textit{my previous} preference");
(iii) Intent Preservation: The query must remain semantically consistent with the execution of $C$.

To ensure that each generated query is genuinely memory-dependent, we first filter out samples that reveal parameter values through explicit mentions or implicit hints.  We then introduce a discriminator LLM, which attempts to reconstruct the correct tool invocation $C$ using only the query $Q$ and the tool’s API documentation, without access to historical memory. A sample is retained only if the discriminator fails, guaranteeing that correct tool invocation is impossible without retrieving relevant memory. Details are provided in Appendix \ref{app:quality_control}.

We employ \textbf{Qwen3-Next-80B-A3B-Instruct} \footnote{\url{https://huggingface.co/Qwen/Qwen3-Next-80B-A3B-Instruct}} as the backbone LLM for target tool selection and parameter grounding. For the reverse implicit query generation stage, we employ \textbf{Kimi-K2-Thinking}\footnote{\url{https://huggingface.co/moonshotai/Kimi-K2-Thinking}}. Finally, we generate a total of \textbf{400} memory-dependent tool-use queries grounded in \textbf{2,029} long conversational sessions, with an average of \textbf{13} turns per session.
These samples constitute the final \textsc{Mem2ActBench} dataset used in all subsequent experiments.

\subsection{Task Formalization}

We define the evaluation task as a conditional generation problem. Given a memory sequence $\mathcal{M}$ and a user query $q$, the agent generates the optimal tool invocation $\hat{c}$ by maximizing:

\begin{equation}
\hat{c} = \arg\max_{c \in \mathcal{C}} P(c \mid q, \mathcal{M})
\end{equation}

where $c$ consists of a selected tool $T$ and parameter values ${v_p}$. Each $v_p$ is derived by reasoning over the context:

\begin{equation}
v_p = f_{\theta}(p, q, \mathcal{M})
\end{equation}

subject to the constraint that $v_p$ is strictly grounded in $\mathcal{M}$.

\subsection{Human Verification}
We conduct expert verification to assess the reliability of \textsc{Mem2ActBench} at three critical stages: fact extraction, conflict resolution, and memory-dependent task formulation. A total of five expert annotators, each holding advanced degrees in fields such as Computational Linguistics, Computer Science, or Artificial Intelligence, and with prior experience in evaluating AI models, were recruited. Each item was independently reviewed by at least two annotators, ensuring thorough evaluation. Disagreements between annotators were resolved through discussion.

For fact extraction, annotators judge whether each fact is (i) entailed by the dialogue context and (ii) correctly normalized. For conflict resolution, they assess whether the resulting memory evolution chain is coherent and logically consistent. For memory dependency, annotators determine whether the gold tool invocation remains underdetermined given the user query alone (i.e., would be infeasible to infer without access to long-term memory). As shown in Table~\ref{tab:human_verification}, we obtain high validation rates across all stages, indicating that \textsc{Mem2ActBench} provides faithful memory states and that its tool-use tasks intrinsically require memory rather than surface-level reasoning.

\begin{table}[t]
\centering
\resizebox{\columnwidth}{!}{
\begin{tabular}{lcc}
\toprule
\textbf{Aspect} & \textbf{\#Samples} & \textbf{Validated (\%)} \\
\midrule
Fact Extraction Accuracy      & 200 & 96.5 \\
Conflict Resolution Quality   & 150 & 86.7 \\
Memory Dependency Validity    & 200 & 91.3 \\
\bottomrule
\end{tabular}}
\caption{Expert verification on randomly sampled instances from three stages of the \textsc{Mem2ActBench} pipeline.}
\label{tab:human_verification}
\end{table}

\section{Experiment}
\label{sec:experiment}

\subsection{Experimental Setup}
\paragraph{Datasets.}
We selected only the conversation histories that contain all the necessary QA evidence, ensuring the original order is preserved, with a total of 429  sessions used.

\paragraph{Baselines.}
We evaluate the following representative agent memory systems on \textsc{Mem2ActBench}: Long-term Memory (RAG), Generative Agents \cite{park2023generative}, SCM\cite{wang2023scm}, Langmem\cite{langmem}, MemTree\cite{rezazadeh2024isolated}, Mem0\cite{chhikara2025mem0buildingproductionreadyai} and A-Mem\cite{xu2025amemagenticmemoryllm}. To control for backbone model capacity, we conduct experiments using three model scales of the Qwen2.5 family, namely \textbf{Qwen2.5-7B-Instruct}, \textbf{Qwen2.5-32B-Instruct}, and \textbf{Qwen2.5-72B-Instruct}\cite{qwen2.5}, as the inference backbone for all memory systems. All models are evaluated under fixed decoding settings (temperature = 0.0) to ensure result stability and comparability across scales. For memory systems involving retrieval, we use \textbf{BGE-m3}\cite{chen2024m3} as the embedding model.

\paragraph{Evaluation Metrics.}
We evaluate memory-based tool-call generation with three metrics: \textbf{F1} for parameter-level precision/recall, \textbf{BLEU-1} for unigram overlap with the reference, and \textbf{Tool Accuracy (TA)}, which is True only if the correct tool is used and all parameters match exactly. In the main results, we provide the ground-truth tool to control for tool-selection errors and focuses the comparison on memory-based parameter grounding.

\begin{table*}
\centering
\resizebox{\textwidth}{!}{
    \begin{tabular}{l ccc cc ccc ccc ccc}
    \toprule
    \multirow{2}{*}{\textbf{Method}} & 
    \multicolumn{3}{c}{\textit{Qwen2.5-72B-Instruct}} & 
    \multicolumn{3}{c}{\textit{Qwen2.5-32B-Instruct}} & 
    \multicolumn{3}{c}{\textit{Qwen2.5-7B-Instruct}} & 
    \multicolumn{3}{c}{\textbf{Average}} \\
    
    \cmidrule(lr){2-4} \cmidrule(lr){5-7} \cmidrule(lr){8-10} \cmidrule(lr){11-13}
    
     & \textbf{F1} & \textbf{BLEU} & \textbf{TA}  & \textbf{F1} & \textbf{BLEU}  & \textbf{TA}   & \textbf{F1} & \textbf{BLEU}  & \textbf{TA} & \textbf{F1} & \textbf{BLEU}  & \textbf{TA} \\
    \midrule
    
    LTMemory & 35.32 & 67.93 & 92.00 & \textbf{33.87} & 67.71 & 90.20 & 26.71 & 64.07 & 87.25 & 31.97 & 66.57 & 89.82 \\ 
    SCM\cite{wang2023scm} & 22.73 & 62.04 & 96.25 & 17.35 & 59.37 & 95.75 & 14.99 & 57.37 & 91.00 & 18.36 & 59.59 & 94.33 \\ 
    Generative Agents\cite{park2023generative} & 22.38 & 62.58 & 95.50 & 17.81 & 59.76 & 96.24 & 14.44 & 55.59 & 89.75 & 18.21 & 59.31 & 93.83 \\ 
    MemTree\cite{rezazadeh2024isolated} & 33.21 & \textbf{68.17} & 94.25 & 31.89 & 67.69 & 94.50 & 24.60 & 63.91 & 88.50 & 29.90 & 66.59 & 92.42 \\ 
    Mem0\cite{chhikara2025mem0buildingproductionreadyai} & 28.95 & 66.47 & \textbf{96.75} & 24.52 & 64.36 & 97.00 & 14.21 & 57.37 & 93.97 & 22.56 & 62.73 & 95.91 \\ 
    Langmem\cite{langmem} & 24.01 & 63.06 & 96.25 & 18.72 & 58.27 & \textbf{97.25} & 17.06 & 58.93 & \textbf{96.50} & 19.93 & 60.09 & \textbf{96.67} \\ 
    A-mem\cite{xu2025amemagenticmemoryllm}& \textbf{35.93} & 67.93 & 93.25 & 33.72 & \textbf{67.92} & 92.25 & \textbf{30.99} & \textbf{66.47} & 94.75 & \textbf{33.55} & \textbf{67.44} & 93.42 \\ 
    \bottomrule
    \end{tabular}
    }
    \caption{Experimental results for different memory methods across multiple model sizes.}
    \label{tab:main_results}

\end{table*}

\subsection{Main Results}

Table\ref{tab:main_results} suggests that TA stays tightly clustered ($\sim$87--97\%) and changes little from 32B to 72B on average, indicating that most remaining errors are not structural but semantic. In contrast, argument grounding shows clear headroom: mean F1 increases from 20.4 (7B) to 28.9 (72B), with diminishing returns beyond 32B ($\approx +3.5$ F1 from 32B$\rightarrow$72B), implying that scaling mainly improves post-retrieval composition. The method ranking reveals that A-mem and LTMemory form the top cluster (72B F1$=$35.9/35.3) and nearly converge at scale, while MemoryTree remains competitive (33.2) but retains a $\sim$2--3 F1 gap, suggesting structured memory helps but does not fully resolve parameter assembly. Notably, the largest scaling gain appears in weaker memory managers (e.g., Mem0: $+14.7$ F1 from 7B$\rightarrow$72B), consistent with larger models compensating via stronger cross-turn inference when memory organization is suboptimal.

\section{Discussions}
\subsection{Retriever Analysis}

To probe whether performance is mainly limited by memory retrieval rather than reasoning, we conduct a controlled comparison across three retrieval conditions (no retrieval, passive retrieval with standard retrievers, and oracle retrieval with ground-truth memories. As shown in Table~\ref{tab:retrieval_ablation}, the best passive retrieval result is achieved by the hybrid retriever at $k{=}5$, reaching $F1 \approx 30.7$. In contrast, oracle retrieval boosts performance to $F1 \approx 53.8$, creating a gap of over 23 F1 points. This large margin suggests that the dominant bottleneck is evidence hitting/retrieval quality, rather than the model’s pure reasoning once the correct supporting memories are available. This finding aligns with our benchmark goal of evaluating memory application for parameter grounding under underspecified requests: improving performance require stronger evidence-hitting mechanisms (e.g., better indexing, retriever training, and query formulation) rather than simply scaling the backbone model.

\definecolor{grayTop1}{gray}{0.95}  
\definecolor{grayTop5}{gray}{0.85}  
\definecolor{grayTop10}{gray}{0.75} 

\begin{table}[t]
    \centering
    \small
    \caption{Performance comparison under varying top-$k$ retrieval settings.
    Shading in the \textbf{Recall@k} column indicates retrieval depth (darker denotes more retrieved documents).
    Best results among passive retrieval methods are highlighted in bold.}
    \label{tab:retrieval_ablation}
    \resizebox{\columnwidth}{!}{
    \begin{tabular}{lcccc}
        \toprule
        \textbf{Retrieval Strategy} & \textbf{Recall@k} & \textbf{F1} & \textbf{BLEU} & \textbf{TSA} \\
        \midrule
        
        \textit{No Retrieval} & -- & 10.0 & 8.9 & 73.8 \\
        
        \midrule
        \multicolumn{5}{l}{\textit{Passive Retrieval}} \\
        
        \multirow{3}{*}{BM25} 
          & \cellcolor{grayTop1} 1   & 29.0 & 28.0 & 87.0 \\
          & \cellcolor{grayTop5} 5   & 26.9 & 26.2 & 90.2 \\
          & \cellcolor{grayTop10} 10 & 27.6 & 26.7 & 87.5 \\
        \addlinespace
        
        \multirow{3}{*}{Dense} 
          & \cellcolor{grayTop1} 1   & 25.6 & 25.3 & 83.0 \\
          & \cellcolor{grayTop5} 5   & 29.9 & 29.1 & 90.0 \\
          & \cellcolor{grayTop10} 10 & 28.7 & 28.3 & 88.8 \\
        \addlinespace
        
        \multirow{3}{*}{Hybrid} 
          & \cellcolor{grayTop1} 1   & 24.9 & 24.3 & 85.0 \\
          & \cellcolor{grayTop5} 5   
          & \textbf{30.7} & \textbf{29.7} & 86.0 \\
          & \cellcolor{grayTop10} 10 & 30.3 & 29.5 & 88.8 \\
        
        \midrule
        \multicolumn{5}{l}{\textit{Perfect Retrieval}} \\
        Oracle & -- & \textbf{53.8} & \textbf{53.7} & 88.2 \\
        
        \bottomrule
    \end{tabular}
    }
\end{table}

\subsection{Impact of Memory Distance}

While the main results demonstrate the overall capability of memory models in tool-use tasks, a critical question remains: does the physical distance between the relevant memory and the current query affect the model's reasoning performance? Existing research on long-context LLMs suggests a "lost-in-the-middle" phenomenon or performance degradation as key information recedes further into the history. To investigate this in the context of \textsc{Mem2ActBench}, we conducted a fine-grained analysis of model performance relative to the "memory distance". For each sample, we identify the earliest turn that provides evidence for any required tool parameter, denoted $t_{\text{earliest}}$, within a conversation of length $L$. We use the normalized position $P_{\text{mem}} = t_{\text{earliest}}/L$ and bucket samples into four quartiles (0--25\%, 25--50\%, 50--75\%, 75--100\%).

\begin{figure}
    \centering
    \includegraphics[width=1\linewidth]{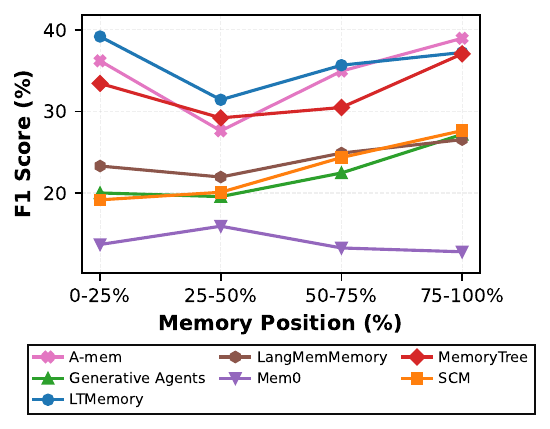}
    \caption{F1 score versus the normalized position of the earliest supporting memory.}
    \label{fig:distance}
\end{figure}

\paragraph{Findings.}

Figure~\ref{fig:distance} reveals a pronounced positional bias: most baselines achieve higher F1 when the supporting evidence appears in the early (0--25\%) or recent (75--100\%) context, but drop sharply when it lies in the mid-history (25--50\%), forming a clear mid-context valley. AgenticMemory falls from $\sim$36\% to $\sim$25\% in the 25--50\% bin, echoing a ``lost-in-the-middle''-like failure in tool-use. By contrast, LTMemory is more position-robust, with F1 remaining above 30\% across bins. These results directly support that even with retrieval, long-term memory is not reliably \emph{applied} for parameter grounding when evidence is buried in the middle of lengthy interaction histories, leaving mid-context usage as a key bottleneck for memory-centric tool agents.

\subsection{Parameter Grounding and Complexity}

To figure out where argument grounding fails, we report \textbf{Slot Accuracy}, as the exact match of each individual argument value, and break it down by grounding type and value complexity.

\paragraph{Grounding Type Analysis.}
We categorize parameters by how their values are supported in the dialogue history: \textbf{Explicit} (directly stated, e.g., ``New York''), \textbf{Inferred} (needs a semantic conversion, e.g., ``upcoming week'' $\rightarrow$ \texttt{days=7}), and \textbf{Default} (not mentioned and should be filled from the tool schema).

\begin{figure}[t]
    \centering
    \includegraphics[width=1\linewidth]{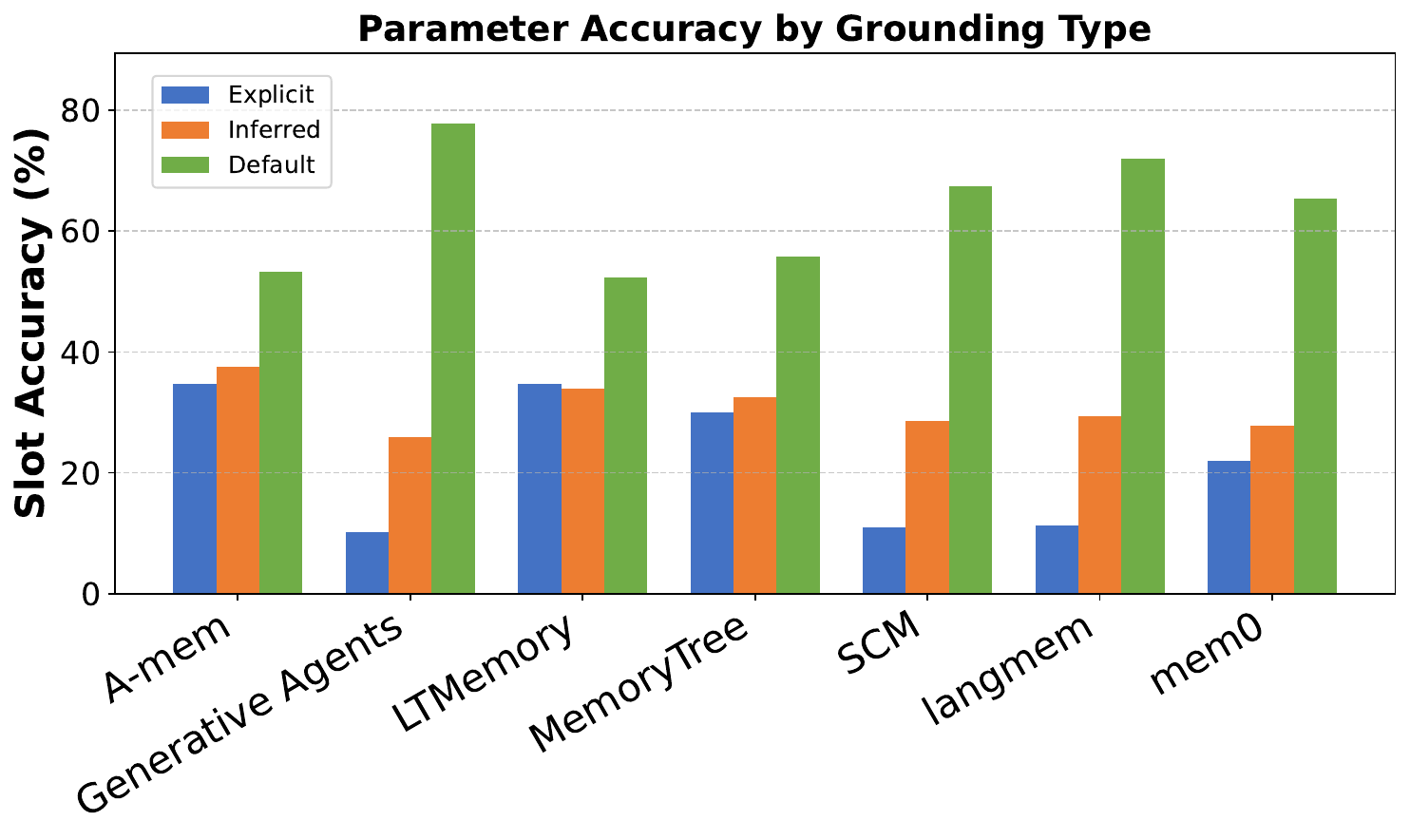}
    \includegraphics[width=1\linewidth]{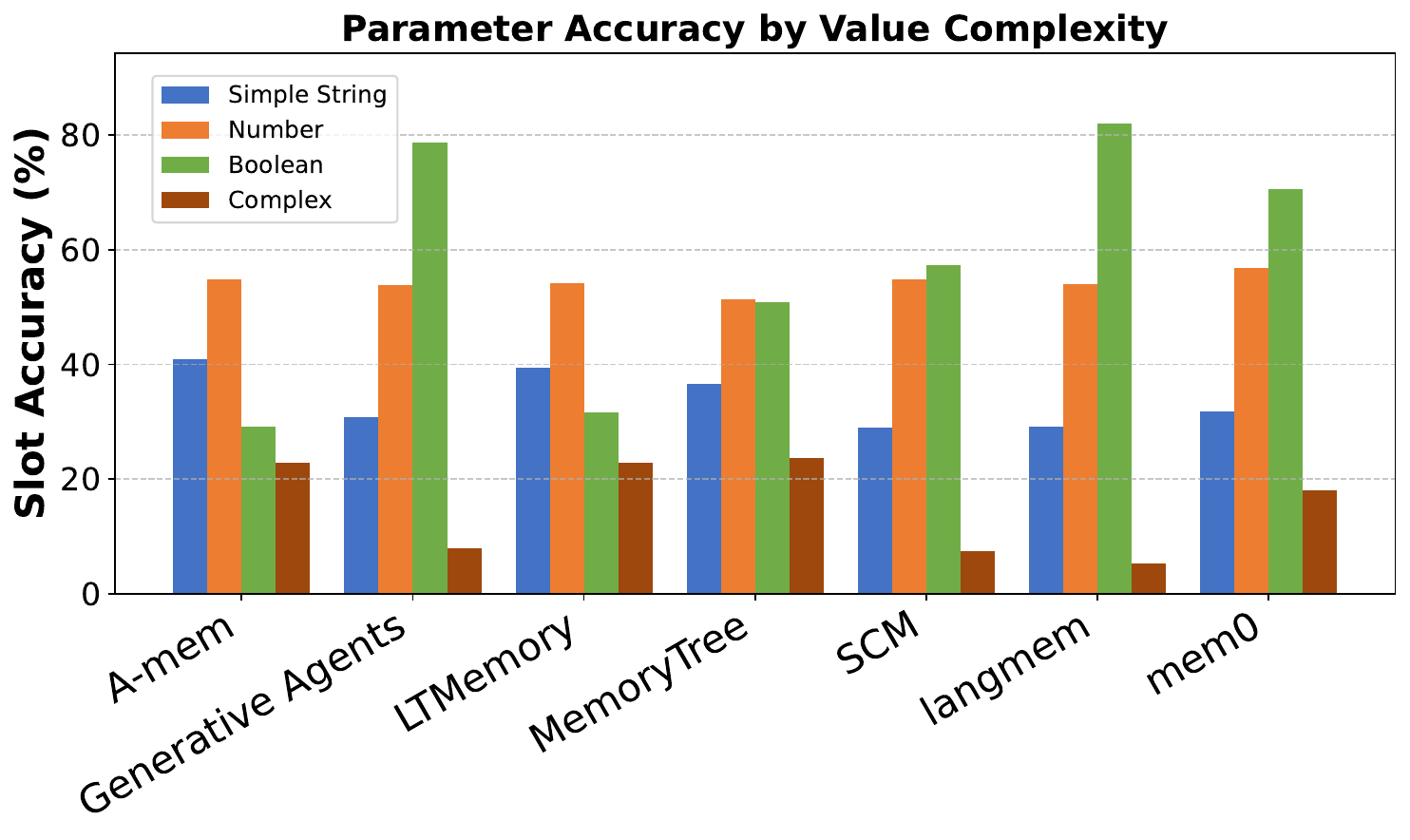}
    \caption{Breakdown of Slot Accuracy by Value Complexity (top) and Grounding Type (bottom).}
    \label{fig:parameter}
\end{figure}

For 72B-scale models, Explicit and Inferred show a small gap, indicating that once the right evidence is retrieved, semantic transformation is not the main difficulty. The largest errors come from Default values: models often fail to notice that a value was never specified, and instead generate a plausible default, sometimes guided by distractors in long histories (Figure~\ref{fig:parameter}).

\paragraph{Value Complexity.}
We further bucket values as \textbf{Simple String} ($\leq$30 chars), \textbf{Number}, \textbf{Boolean}, and \textbf{Complex} (long strings, specialized identifiers such as URLs/addresses, or nested structures).

Slot Accuracy decreases as values become more complex. Models handle Simple Strings and Numbers reasonably well, but performance drops sharply on \textbf{Complex} values, which points to weak lossless retention (e.g., truncation or character-level corruption of identifiers). Boolean accuracy also varies across frameworks, indicating that it depends on both the grounding context and how each system enforces tool constraints (Figure~\ref{fig:parameter}).

\subsection{Tool Selection Robustness}

We stress-test tool choice by increasing the candidate set to $N\in\{1,2,5\}$, where each query is paired with $N\!-\!1$ distractor tools. Distractors are sampled either \textbf{randomly} (uniformly from the tool library) or as \textbf{hard negatives} (distractor tools most semantically similar to the ground-truth tool), which tests fine-grained intent separation (e.g., \textit{search} vs.\ \textit{book}). We report Tool Selection Accuracy (TSA) and end-to-end Exact Match (EM) (correct tool and all arguments). For diagnosis, we also report Arg\_F1 conditioned on selecting the correct tool.

Table~\ref{tab:tool-selection-robustness} shows that with random negatives, TSA remains high and nearly unchanged (93.50--95.50\%) as $N$ increases, suggesting that models handle unrelated distractors well. In contrast, hard negatives cause a steep drop in TSA, from 94.50\% ($N{=}1$) to 69.75\% ($N{=}5$), indicating difficulty when tool semantics overlap. EM stays low in all settings (14.25--18.25\%), even when TSA is above 93\%, which suggests that argument grounding is the main bottleneck. This is also reflected in Arg\_F1: under hard negatives (given the correct tool), it decreases from 29.88 to 22.64, implying that similar distractors can also hurt parameter extraction and inference.

\begin{table}[t]
\centering
\small
\resizebox{0.9\columnwidth}{!}{
\begin{tabular}{lccc}
\toprule
\textbf{Candidate size $N$} & \textbf{1} & \textbf{2} & \textbf{5} \\
\midrule
\multicolumn{4}{l}{\textit{Random negatives}} \\
TSA ($\uparrow$)        & 94.50 & 95.50 & 93.50 \\
EM ($\uparrow$)         & 18.25 & 18.00 & 17.00  \\
Arg\_F1 ($\uparrow$)     & 29.88 & 29.98 & 28.27 \\
\midrule
\multicolumn{4}{l}{\textit{Hard negatives}} \\
TSA ($\uparrow$)        & 94.50 & 78.00 & 69.75 \\
EM ($\uparrow$)         & 18.25 & 16.50 & 14.25 \\
Arg\_F1($\uparrow$)      & 29.88 & 27.17 & 22.64  \\
\bottomrule
\end{tabular}
}
\caption{Tool selection robustness under different candidate tool set sizes ($N\in\{1,2,5\}$).}
\label{tab:tool-selection-robustness}
\end{table}

\subsection{Error Mode Diagnosis}
\label{sec:error_diagnosis}

To characterize why agents fail on memory-grounded parameter filling, we conduct a fine-grained attribution analysis over erroneous predictions. 
We categorize failures into five types:
\textit{(i) Retrieval Miss}, required evidence is absent from retrieved context;
\textit{(ii) Retrieved-but-Unused}, evidence is retrieved but not utilized;
\textit{(iii) Hallucinated Default}, schema defaults are incorrectly overridden or fabricated;
\textit{(iv) Lossless Retention Failure}, long/structured values are corrupted (e.g., truncation or character-level errors);
and \textit{(v) Tool Selection Error}, an incorrect tool is selected.

\begin{figure}
    \centering
    \includegraphics[width=1\linewidth]{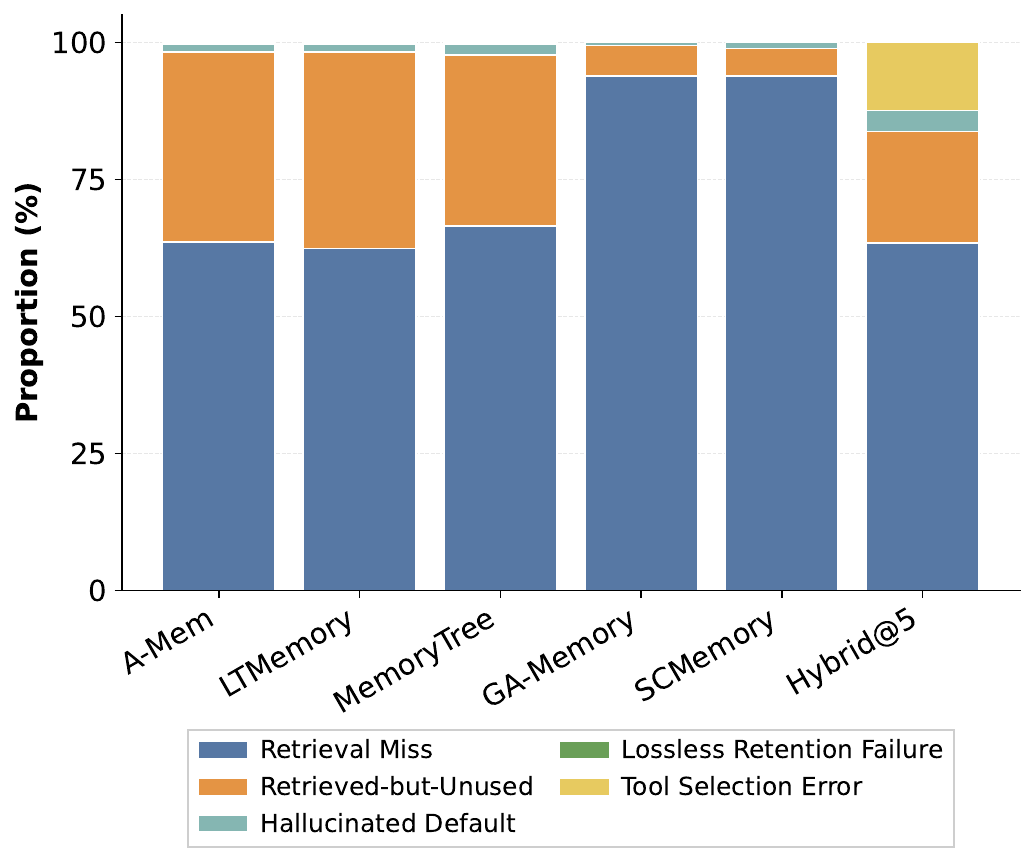}
     \caption{Distribution of failure modes across different memory frameworks.}
   \label{fig:error_mode}
\end{figure}

\paragraph{Findings.}
Figure \ref{fig:error_mode} shows that: 
(i) As memory frameworks become stronger, failures shift from \emph{accessibility} (retrieval misses dominating weak baselines) to \emph{post-retrieval reasoning} (retrieved-but-unused rising substantially), suggesting retrieval alone is insufficient.
(ii) Tool selection is largely robust: agentic frameworks exhibit negligible tool selection errors, while a passive retrieval baseline shows a noticeable fraction of tool-selection failures.
(iii) Retrieval Miss remains the largest category even for strong systems, highlighting the enduring difficulty of locating sparse but critical evidence under implicit queries.

\section{Conclusion}
In this paper, we introduce \textsc{Mem2ActBench} for evaluating whether tool-augmented agents can effectively apply long-term memory to drive task execution. We construct a benchmark through an automated pipeline that simulates real-world interrupted interactions, generating memory-dependent tool calls. Our experiments reveal a significant gap in current memory frameworks, particularly in parameter grounding with mid-context memories often overlooked. These results highlight the limitations of current systems in proactively utilizing long-term memory, especially when tasks are underspecified. Future research should focus on enhancing the active utilization of memory, particularly in scenarios that require reasoning over dispersed, incomplete information.

\section*{Limitations}
\textsc{Mem2ActBench} is designed to evaluate memory-grounded parameterization in tool-based tasks under controlled conditions. However, it is limited to offline tool-call generation, using a fixed backbone model family, which does not reflect the diversity of real-world models. The benchmark also excludes interactive execution settings, where agents adapt to feedback over time. Additionally, while automated task generation helps scale the dataset, it may not fully capture the complexities of real-world dialogues. Lastly, human verification introduces potential biases in the validation process, especially in edge cases.

\section*{Ethical considerations}
\textsc{Mem2ActBench} is constructed by synthesizing interaction histories from publicly available datasets, including task-oriented tool-use data from ToolACE and BFCL, and conversational content from OpenAssistant (OASST1). No new data were collected from end users or through human-subject experiments. Following the ethical practices described by these source datasets and the ACL ethics guidance, we take steps to reduce privacy risks by releasing only processed benchmark instances necessary for evaluating memory-grounded tool use, and by applying automated redaction or rewriting to remove obvious personally identifiable information (e.g., emails, phone numbers, account identifiers) when encountered.

\bibliography{custom}

@article{liu-etal-2024-lost,
    title = "Lost in the Middle: How Language Models Use Long Contexts",
    author = "Liu, Nelson F.  and
      Lin, Kevin  and
      Hewitt, John  and
      Paranjape, Ashwin  and
      Bevilacqua, Michele  and
      Petroni, Fabio  and
      Liang, Percy",
    journal = "Transactions of the Association for Computational Linguistics",
    volume = "12",
    year = "2024",
    address = "Cambridge, MA",
    publisher = "MIT Press",
    url = "https://aclanthology.org/2024.tacl-1.9/",
    doi = "10.1162/tacl_a_00638",
    pages = "157--173"
}

@inproceedings{ICLR2025_663865ea,
 author = {Liu, Weiwen and Huang, Xu and Zeng, Xingshan and hao, xinlong and Yu, Shuai and Li, Dexun and Wang, Shuai and Gan, Weinan and Liu, Zhengying and Yu, Yuanqing and WANG, Zezhong and Wang, Yuxian and Ning, Wu and Hou, Yutai and Wang, Bin and Wu, Chuhan and Xinzhi, Wang and Liu, Yong and Wang, Yasheng and Tang, Duyu and Tu, Dandan and Shang, Lifeng and Jiang, Xin and Tang, Ruiming and Lian, Defu and Liu, Qun and Chen, Enhong},
 booktitle = {International Conference on Representation Learning},
 editor = {Y. Yue and A. Garg and N. Peng and F. Sha and R. Yu},
 pages = {41359--41381},
 title = {ToolACE: Winning the Points of LLM Function Calling},
 url = {https://proceedings.iclr.cc/paper_files/paper/2025/file/663865ea167425c6c562cb0b6bcf76c7-Paper-Conference.pdf},
 volume = {2025},
 year = {2025}
}

@misc{modarressi2024retllmgeneralreadwritememory,
      title={RET-LLM: Towards a General Read-Write Memory for Large Language Models}, 
      author={Ali Modarressi and Ayyoob Imani and Mohsen Fayyaz and Hinrich Schütze},
      year={2024},
      eprint={2305.14322},
      archivePrefix={arXiv},
      primaryClass={cs.CL},
      url={https://arxiv.org/abs/2305.14322}, 
}

@inproceedings{zhong2024memorybank,
  title={Memorybank: Enhancing large language models with long-term memory},
  author={Zhong, Wanjun and Guo, Lianghong and Gao, Qiqi and Ye, He and Wang, Yanlin},
  booktitle={Proceedings of the AAAI Conference on Artificial Intelligence},
  volume={38},
  number={17},
  pages={19724--19731},
  year={2024}
}

@inproceedings{park2023generative,
  title={Generative agents: Interactive simulacra of human behavior},
  author={Park, Joon Sung and O'Brien, Joseph and Cai, Carrie Jun and Morris, Meredith Ringel and Liang, Percy and Bernstein, Michael S},
  booktitle={Proceedings of the 36th annual acm symposium on user interface software and technology},
  pages={1--22},
  year={2023}
}

@misc{packer2024memgptllmsoperatingsystems,
      title={MemGPT: Towards LLMs as Operating Systems}, 
      author={Charles Packer and Sarah Wooders and Kevin Lin and Vivian Fang and Shishir G. Patil and Ion Stoica and Joseph E. Gonzalez},
      year={2024},
      eprint={2310.08560},
      archivePrefix={arXiv},
      primaryClass={cs.AI},
      url={https://arxiv.org/abs/2310.08560}, 
}

@misc{xu2025amemagenticmemoryllm,
      title={A-MEM: Agentic Memory for LLM Agents}, 
      author={Wujiang Xu and Zujie Liang and Kai Mei and Hang Gao and Juntao Tan and Yongfeng Zhang},
      year={2025},
      eprint={2502.12110},
      archivePrefix={arXiv},
      primaryClass={cs.CL},
      url={https://arxiv.org/abs/2502.12110}, 
}

@inproceedings{yamashita2023realpersonachat,
  title={RealPersonaChat: A realistic persona chat corpus with interlocutors’ own personalities},
  author={Yamashita, Sanae and Inoue, Koji and Guo, Ao and Mochizuki, Shota and Kawahara, Tatsuya and Higashinaka, Ryuichiro},
  booktitle={Proceedings of the 37th Pacific Asia Conference on Language, Information and Computation},
  pages={852--861},
  year={2023}
}

@inproceedings{xu2022beyond,
  title={Beyond goldfish memory: Long-term open-domain conversation},
  author={Xu, Jing and Szlam, Arthur and Weston, Jason},
  booktitle={Proceedings of the 60th annual meeting of the association for computational linguistics (volume 1: long papers)},
  pages={5180--5197},
  year={2022}
}

@article{maharana2024evaluating,
  title={Evaluating very long-term conversational memory of llm agents},
  author={Maharana, Adyasha and Lee, Dong-Ho and Tulyakov, Sergey and Bansal, Mohit and Barbieri, Francesco and Fang, Yuwei},
  journal={arXiv preprint arXiv:2402.17753},
  year={2024}
}

@article{wu2024longmemeval,
  title={Longmemeval: Benchmarking chat assistants on long-term interactive memory},
  author={Wu, Di and Wang, Hongwei and Yu, Wenhao and Zhang, Yuwei and Chang, Kai-Wei and Yu, Dong},
  journal={arXiv preprint arXiv:2410.10813},
  year={2024}
}

@article{hu2025evaluating,
  title={Evaluating memory in llm agents via incremental multi-turn interactions},
  author={Hu, Yuanzhe and Wang, Yu and McAuley, Julian},
  journal={arXiv preprint arXiv:2507.05257},
  year={2025}
}

@inproceedings{patil2025bfcl,
  title={The Berkeley Function Calling Leaderboard (BFCL): From Tool Use to Agentic Evaluation of Large Language Models},
  author={Patil, Shishir G. and Mao, Huanzhi and Cheng-Jie Ji, Charlie and Yan, Fanjia and Suresh, Vishnu and Stoica, Ion and E. Gonzalez, Joseph},
  booktitle={Forty-second International Conference on Machine Learning},
  year={2025},
}

@article{grootendorst2022bertopic,
  title={BERTopic: Neural topic modeling with a class-based TF-IDF procedure},
  author={Grootendorst, Maarten},
  journal={arXiv preprint arXiv:2203.05794},
  year={2022}
}

@misc{qwen2.5,
    title = {Qwen2.5: A Party of Foundation Models},
    url = {https://qwenlm.github.io/blog/qwen2.5/},
    author = {Qwen Team},
    month = {September},
    year = {2024}
}

@inproceedings{chen2024m3,
  title={M3-embedding: Multi-linguality, multi-functionality, multi-granularity text embeddings through self-knowledge distillation},
  author={Chen, Jianlyu and Xiao, Shitao and Zhang, Peitian and Luo, Kun and Lian, Defu and Liu, Zheng},
  booktitle={Findings of the Association for Computational Linguistics ACL 2024},
  pages={2318--2335},
  year={2024}
}

@article{kopf2023openassistant,
  title={Openassistant conversations-democratizing large language model alignment},
  author={K{\"o}pf, Andreas and Kilcher, Yannic and Von R{\"u}tte, Dimitri and Anagnostidis, Sotiris and Tam, Zhi Rui and Stevens, Keith and Barhoum, Abdullah and Nguyen, Duc and Stanley, Oliver and Nagyfi, Rich{\'a}rd and others},
  journal={Advances in neural information processing systems},
  volume={36},
  pages={47669--47681},
  year={2023}
}

@article{wang2023scm,
  title={Scm: Enhancing large language model with self-controlled memory framework},
  author={Wang, Bing and Liang, Xinnian and Yang, Jian and Huang, Hui and Wu, Shuangzhi and Wu, Peihao and Lu, Lu and Ma, Zejun and Li, Zhoujun},
  journal={arXiv e-prints},
  pages={arXiv--2304},
  year={2023}
}

@article{rezazadeh2024isolated,
  title={From isolated conversations to hierarchical schemas: Dynamic tree memory representation for llms},
  author={Rezazadeh, Alireza and Li, Zichao and Wei, Wei and Bao, Yujia},
  journal={arXiv preprint arXiv:2410.14052},
  year={2024}
}

@misc{chhikara2025mem0buildingproductionreadyai,
      title={Mem0: Building Production-Ready AI Agents with Scalable Long-Term Memory}, 
      author={Prateek Chhikara and Dev Khant and Saket Aryan and Taranjeet Singh and Deshraj Yadav},
      year={2025},
      eprint={2504.19413},
      archivePrefix={arXiv},
      primaryClass={cs.CL},
      url={https://arxiv.org/abs/2504.19413}, 
}

@misc{langmem,
  author       = {{LangChain}},
  title        = {LangMem (langchain-ai/langmem)},
  year         = {2025},
  month        = oct,
  note         = {Version 0.0.30; accessed 2026-01-03},
  howpublished = {\url{https://github.com/langchain-ai/langmem}}
}

@misc{kim2025dialsimdialoguesimulatorevaluating,
      title={DialSim: A Dialogue Simulator for Evaluating Long-Term Multi-Party Dialogue Understanding of Conversational Agents}, 
      author={Jiho Kim and Woosog Chay and Hyeonji Hwang and Daeun Kyung and Hyunseung Chung and Eunbyeol Cho and Yeonsu Kwon and Yohan Jo and Edward Choi},
      year={2025},
      eprint={2406.13144},
      archivePrefix={arXiv},
      primaryClass={cs.CL},
      url={https://arxiv.org/abs/2406.13144}, 
}

\appendix
\section{Data Processing Details}
\label{sec:data_process_details}

\subsection{Pipeline Implementation}
We employ a standardized pipeline to unify heterogeneous data sources into a consistent multi-turn format. The processing specifics for each source are as follows:

\begin{itemize}
    \item \textbf{ToolACE Processing:} We parse raw interaction traces using a custom stack-based algorithm to handle nested bracket structures (e.g., \texttt{[Function(args...)]}). These parsed traces are then refined into natural language dialogues using \texttt{Qwen/Qwen3-Next-80B-A3B-Instruct} (temperature=0.0) to ensure conversational fluidity while preserving execution logic.
    
    \item \textbf{BFCL v3 Synthesis:} To transform static query-response pairs into dynamic interactions, we utilize the same LLM engine (temperature=0.0) to synthesize multi-round histories. This involves expanding single-turn ground truths into coherent contexts containing user clarifications and sequential tool invocations.
    
    \item \textbf{OASST1 Formatting:} We reconstruct full conversation threads by tracing leaf nodes to the root, filtering for high-quality responses (\texttt{rank=0}). The data is further processed by deduplicating based on the longest conversation path per prompt and translating non-English samples to English via the Google Translate API to maintain linguistic consistency.
\end{itemize}

\paragraph{Data Schema. }
All processed data is serialized into a unified JSONL format compatible with standard chat completion APIs. Table~\ref{tab:json_sample} illustrates a representative sample structure.

\begin{table}[t]
\caption{Unified JSON schema used for training, aligned with standard tool-use formats.}
\label{tab:json_sample}
\centering
\begin{minipage}{\columnwidth}
\begin{Verbatim}[
  fontsize=\footnotesize,
  breaklines=true,
  breakanywhere=true,
  breaksymbolleft={}
]
{
  "id": "toolace_sample_01",
  "tools": [
    {
      "type": "function",
      "function": {
        "name": "search_api",
        "description": "Search for information online.",
        "parameters": { ... }
      }
    }
  ],
  "conversation_history": [
    {
      "role": "user",
      "content": "Check the weather in NY."
    },
    {
      "role": "assistant",
      "content": "I will check the forecast for New York.",
      "tool_calls": [
        {
          "id": "call_abc123",
          "type": "function",
          "function": {
            "name": "weather_api",
            "arguments": "{\"location\": \"New York, NY\"}"
          }
        }
      ]
    },
    {
      "role": "tool",
      "tool_call_id": "call_abc123",
      "name": "weather_api",
      "content": "{\"temp\": \"20C\", \"condition\": \"Sunny\"}"
    }
  ]
}
\end{Verbatim}
\end{minipage}
\end{table}

\subsection{Fact Extraction, Semantic Clustering, and Local Conflict Resolution}
\label{app:data_processing_details}

Our pipeline transforms raw dialogue sessions into a coherent memory structure through three sequential stages: extracting atomic facts, clustering them by semantic topic, and resolving inconsistencies within each local group.

\paragraph{Fact Extraction.}
We employ LLM to process each dialogue session and extract structured facts formatted as triplets: \texttt{(attribute, fact, source\_id)}. The \texttt{attribute} functions as a normalized category label (e.g., ``Dietary Preference''), while the \texttt{fact} encapsulates the specific atomic statement derived from the user's input. To ensure precision and reproducibility in the extraction process, we set the generation temperature to $0.0$.

\paragraph{Semantic Clustering.}
To unify scattered references to the same topic across disjointed sessions, we utilize \textbf{BERTopic} for semantic aggregation. First, we generate dense vector representations for all extracted attributes using the \textbf{BAAI/bge-m3} embedding model. These embeddings are normalized using the $L_2$ norm to ensure consistent distance measures. For the clustering backend, we employ \textbf{HDBSCAN} to identify semantically related attribute groups. Based on our implementation, we configure the algorithm with a minimum cluster size of 2 (\texttt{min\_cluster\_size=2}) to capture even sparse thematic connections. We utilize the \texttt{euclidean} metric for distance calculation and set the cluster selection method to \texttt{leaf} with an epsilon threshold of $0.01$, favoring finer-grained clusters over broad generalizations. Post-clustering, all attributes within a cluster are mapped to a single canonical representative, ensuring a unified namespace for subsequent processing.

\paragraph{Local Conflict Resolution.}
Once facts are grouped by their canonical attributes, we perform local conflict resolution to establish a consistent timeline for each topic. We prompt LLM to analyze each cluster. The model performs three key tasks:
\begin{enumerate}
    \item \textbf{Chronological Ordering:} It determines the logical sequence of events, producing a sorted list of source IDs (\texttt{sorted\_source\_ids}) that reflects the true evolution of the user's status.
    \item \textbf{Conflict Elimination:} It identifies and explicitly discards source IDs (\texttt{discarded\_source\_ids}) containing obsolete, redundant, or contradictory information that does not fit the coherent narrative.
    \item \textbf{Narrative Synthesis:} It generates a natural language summary and a reasoning trace, explaining how the state evolved (e.g., a change in preference) to facilitate interpretability.
\end{enumerate}
This hierarchical approach ensures that individual topic histories are locally consistent before they are integrated into the global memory evolution chain. Example shown in Tabel \ref{tab:memory_evo_example}. 

\begin{table*}[t]
\centering
\caption{Memory Evolution Example}
\label{tab:memory_evo_example}
\renewcommand{\arraystretch}{1.35}
\setlength{\tabcolsep}{8pt}
\small
\begin{tabular}{p{3cm} p{0.8\textwidth}}
\toprule
\textbf{Item} & \textbf{Content} \\
\midrule

Attribute Group &
Dietary Preference (Vegan) \\

Sorted Source IDs &
\texttt{[oss\_5924, oss\_9685, oss\_8154, toolace\_1099]} \\

Discarded Source IDs &
\texttt{[toolace\_518]} \\

Narrative Summary &
The user has evolved from a vegetarian diet that included croissants and wine to a strictly vegan lifestyle, with a strong preference for vegetables and high-protein meals. All non-vegan foods, including fish, are no longer included. \\

Reasoning &
The preference for fish and vegetables (toolace\_518) conflicts with the current vegan requirement (oss\_5924) and is therefore discarded. The vegetarian status (toolace\_1099) is retained as historical context, reflecting an earlier dietary stage. The current vegan requirement, preference for vegetables, and high-protein intake are retained to represent the complete dietary evolution. \\

Original Facts &
\begingroup
\footnotesize
\setlength{\baselineskip}{0.9\baselineskip}
\ttfamily
\begin{tabular}{@{}p{0.68\textwidth}@{}}
\{ \\
\ \ "toolace\_518": \{ \\
\ \ \ \ "attribute": "Diet Preference (Fish and Vegetables)", \\
\ \ \ \ "fact": "User prefers to include fish and vegetables in their diet" \\
\ \ \}, \\
\ \ "toolace\_1099": \{ \\
\ \ \ \ "attribute": "Dietary Preference (Vegetarian)", \\
\ \ \ \ "fact": "User is vegetarian and enjoys croissants and wine" \\
\ \ \}, \\
\ \ "oss\_5924": \{ \\
\ \ \ \ "attribute": "Dietary Preference (Vegan)", \\
\ \ \ \ "fact": "User requires all lunch options to be vegan" \\
\ \ \}, \\
\ \ "oss\_8154": \{ \\
\ \ \ \ "attribute": "Food Preferences (Vegetables)", \\
\ \ \ \ "fact": "User really likes vegetables" \\
\ \ \}, \\
\ \ "oss\_9685": \{ \\
\ \ \ \ "attribute": "Dietary Preference (Vegetarian with High Protein)", \\
\ \ \ \ "fact": "User is vegetarian and consumes a lot of protein" \\
\ \ \} \\
\}
\end{tabular}
\endgroup
\\

\bottomrule
\end{tabular}
\end{table*}

\subsection{Algorithm for Global Evolution Sequence Construction}
\label{app:algorithm}

In this section, we provide the detailed pseudocode for constructing the Global Evolution Sequence. We employ a modified topological sorting algorithm based on Kahn's algorithm. To handle cyclic dependencies (conflicts) arising from merging heterogeneous data sources, we introduce a deterministic heuristic mechanism.

The core of our conflict resolution strategy lies in the \textit{Cycle Breaking} step. When the topological sort stalls due to a cycle (i.e., no nodes have an in-degree of zero), we explicitly identify the set of deadlocked nodes. From this set, we select a candidate to discard based on the following priority:
\begin{enumerate}
    \item \textbf{Maximum Out-Degree:} We prioritize removing the node with the highest out-degree. A high out-degree implies that the fact imposes ordering constraints on many subsequent facts. Removing it relaxes the graph structure most effectively, allowing the sorting process to resume.
    \item \textbf{Lexicographical Order (Tie-breaker):} If multiple nodes share the same maximum out-degree, we select the one with the lexicographically smallest identifier. This ensures the algorithm is strictly deterministic and reproducible.
\end{enumerate}

The complete procedure is outlined in Algorithm~\ref{alg:global_seq}.

\begin{algorithm}[t]
\caption{Global Evolution Sequence Construction with Deterministic Conflict Resolution}
\label{alg:global_seq}
\small
\raggedright
\begin{algorithmic}[1]
\Require Set of local sequences $\mathcal{S}=\{S_1,S_2,\dots,S_n\}$
\Ensure Globally sorted sequence $G$, set of discarded facts $D$

\State \textbf{Initialize:}
\State Construct directed graph $\mathcal{G}(V,E)$ from $\mathcal{S}$
\State Compute in-degree $deg_{in}(v)$ and out-degree $deg_{out}(v)$ for all $v\in V$
\State $Q \gets \{v\in V \mid deg_{in}(v)=0\}$ \Comment{Initialize queue with source nodes}
\State Sort $Q$ lexicographically \Comment{Ensure determinism}
\State $G \gets []$, $D \gets []$

\While{$|G|+|D|<|V|$}
  \If{$Q$ is not empty}
    \State $u \gets Q.\mathrm{pop}()$
    \State $G.\mathrm{append}(u)$
    \For{each neighbor $v$ of $u$}
      \State $deg_{in}(v) \gets deg_{in}(v)-1$
      \If{$deg_{in}(v)=0$}
        \State $Q.\mathrm{push}(v)$
      \EndIf
    \EndFor
  \Else
    \Comment{\textbf{Cycle detected:} heuristic conflict resolution}
    \State $V_{\text{remain}} \gets V \setminus (G \cup D)$
    \State $v_{\text{drop}} \gets \text{NULL}$, $max\_out \gets -1$
    \For{$v$ in $V_{\text{remain}}$}
      \If{$deg_{out}(v) > max\_out$}
        \State $v_{\text{drop}} \gets v$, $max\_out \gets deg_{out}(v)$
      \ElsIf{$deg_{out}(v)=max\_out$ \textbf{and} $v < v_{\text{drop}}$}
        \State $v_{\text{drop}} \gets v$ \Comment{Lexicographical tie-breaker}
      \EndIf
    \EndFor
    \State $D.\mathrm{append}(v_{\text{drop}})$ \Comment{Discard the conflicting node}
    \For{each neighbor $v$ of $v_{\text{drop}}$}
      \State $deg_{in}(v) \gets deg_{in}(v)-1$
      \If{$deg_{in}(v)=0$}
        \State $Q.\mathrm{push}(v)$
      \EndIf
    \EndFor
  \EndIf
\EndWhile

\State \Return $G, D$
\end{algorithmic}
\end{algorithm}

\section{Quality Control for Reverse Query Generation}
\label{app:quality_control}

To ensure that the generated user queries ($Q$) strictly rely on long-term memory ($M$) to resolve tool parameters ($P$), we implement a filtering pipeline. This process eliminates both surface-level parameter leakage and semantic redundancy where the task is solvable without memory context.

\paragraph{Lexical Leakage Filtering. } 
We first apply a rule-based filter to detect explicit mentions of ground-truth values in $Q$. This check specifically targets parameters derived from memory (marked as \textit{explicit} or \textit{inferred}), ignoring generic schema defaults. The filter rejects $Q$ if it contains: 
(1) \textbf{Exact Matches} of parameter strings (case-insensitive with boundary checks); 
(2) \textbf{Numeric Values} appearing in the query (e.g., price constraints); 
(3) \textbf{Token Overlap} for compound entities (length $>4$), ensuring that distinctive parts of a name (e.g., "California" in "Hotel California") are not leaked; and 
(4) \textbf{Structured Identifiers} (e.g., IDs, emails) detected via substring matching.

\paragraph{Solvability Discriminator. } 
Lexical rules cannot detect semantic leakage (e.g., describing "NYC" as "the Big Apple"). To address this, we employ a \textbf{Blinded LLM Discriminator}. The discriminator is presented with the query $Q$ and tool schema but \textit{denied access} to the memory context $M$. It attempts to predict the tool arguments solely from $Q$. If the discriminator successfully infers the correct parameters, the query is deemed \textit{"Solvable Without Memory"} and rejected. This counterfactual evaluation guarantees that the final samples strictly require memory integration.

\begin{table*}[h]
\centering
\small
\resizebox{\textwidth}{!}{%
\begin{tabular}{@{}p{0.12\textwidth} p{0.22\textwidth} p{0.25\textwidth} p{0.15\textwidth} p{0.15\textwidth}@{}}
\toprule
\textbf{Error Type} & \textbf{Memory Context} & \textbf{Generated Query ($Q$)} & \textbf{Target Param} & \textbf{Verdict} \\ \midrule
\textbf{Direct Leakage} & User wants to visit \textit{Seattle}. & "Help me book a flight to \textbf{Seattle}." & \texttt{city="Seattle"} & \textcolor{red}{Reject} (Rule: Exact Match) \\ \midrule
\textbf{Partial Leakage} & Account ID is \textit{AX-9920}. & "Check my account status, it ends in \textbf{9920}." & \texttt{id="AX-9920"} & \textcolor{red}{Reject} (Rule: Token Split) \\ \midrule
\textbf{Solvable w/o Memory} & User prefers \textit{Italian} food. & "Find me a place that serves \textbf{pasta and pizza}." & \texttt{cuisine="Italian"} & \textcolor{red}{Reject} (Discriminator: Semantic Leak) \\ \midrule
\textbf{Hallucination} & (No mention of time). & "Book the ticket for \textbf{tomorrow morning}." & \texttt{time="09:00"} & \textcolor{red}{Reject} (Discriminator: Invalid Constraint) \\ \midrule
\textbf{Valid Instance} & User mentioned budget is \textit{\$500} in Turn 1. & "Find a flight that fits \textbf{the budget I mentioned earlier}." & \texttt{price=500} & \textcolor{green}{Accept} \\ \bottomrule
\end{tabular}%
}
\caption{Examples of the filtering process. \textbf{Direct/Partial Leakage} is caught by the rule-based filter . \textbf{Solvable/Hallucination} errors are caught by the LLM Discriminator. Only queries that strictly require memory resolution are accepted.}
\label{tab:qc_examples}
\end{table*}

\section{Human Verification Guidelines}
\label{sec:human_verification_guidelines}

To ensure the reliability of \textsc{Mem2ActBench}, a rigorous human verification process was implemented, involving five expert annotators with experience in NLP and agent-evaluation tasks. Each annotator holds advanced degrees in fields such as Computational Linguistics, Computer Science, or Artificial Intelligence, and has prior experience in evaluating AI models. The verification process was divided into three stages: fact extraction, conflict resolution, and memory dependency verification. Annotators cross-checked each item, and disagreements were resolved through discussion to ensure accuracy and consistency.

\subsection{Stage 1 \& 2: Fact Extraction and Conflict Resolution}
Annotators verify whether extracted facts are faithful to the dialogue and whether memory updates (conflicts) are resolved logically. We instruct annotators to strictly distinguish between \textit{updates} (which overwrite old values) and \textit{refinements} (which coexist). Table~\ref{tab:conflict_resolution_cases} illustrates the adjudication logic for edge cases.

\begin{table*}[ht]
\centering
\small
\resizebox{\textwidth}{!}{
\begin{tabular}{@{}p{0.35\textwidth}p{0.35\textwidth}p{0.2\textwidth}@{}}
\toprule
\textbf{Dialogue Context \& Memory State} & \textbf{Pipeline Action / Result} & \textbf{Verdict \& Rationale} \\ \midrule
\textbf{Case: Preference Update} \newline History: ``I prefer aisle seats.'' \newline New Turn: ``Actually, change that to window.'' & \textbf{Action:} \texttt{aisle} $\rightarrow$ \texttt{discarded} \newline \textbf{Result:} \texttt{seat\_pref: window} & \textbf{\textsc{Valid}.} Explicit user correction requires overwriting the previous state (Temporal Obsolescence). \\ \midrule
\textbf{Case: Scope Distinction} \newline History: ``Budget for hotel is \$200.'' \newline New Turn: ``Budget for flight is \$500.'' & \textbf{Action:} \texttt{Merge attempt} \newline \textbf{Result:} \texttt{budget: \$500} & \textbf{\textsc{Invalid}.} Different entity scopes (Hotel vs. Flight). Both facts must be retained as separate entries. \\ \midrule
\textbf{Case: Specificity Refinement} \newline History: ``I want Asian food.'' \newline New Turn: ``Let's go for Sushi.'' & \textbf{Action:} \texttt{Coexist} \newline \textbf{Result:} \texttt{cuisine: Asian}, \texttt{dish: Sushi} & \textbf{\textsc{Valid}.} The new fact refines the old one without contradiction. Both are useful for tool retrieval. \\ \bottomrule
\end{tabular}
}
\caption{Guidelines for validating \textbf{Conflict Resolution}. Annotators must determine if the pipeline correctly identified whether to overwrite, merge, or keep facts based on logical consistency.}
\label{tab:conflict_resolution_cases}
\end{table*}

\subsection{Stage 3: Memory Dependency Verification}

Annotators apply \textit{Information Necessity}: if a competent agent can infer \emph{all} target arguments from the query and the tool schema alone (without consulting long-term memory), the sample is rejected as leakage. In practice, we reject (i) direct/partial leakage that exposes gold values or distinctive substrings, (ii) semantic leakage where the query uniquely identifies the value and becomes solvable without memory, and (iii) unsupported constraints that are grounded in neither memory nor the local query context. Table \ref{tab:qc_examples} provides end-to-end filtering examples from the automated pipeline.

\subsection{Disagreement Resolution Strategy}
We employ a tiered strategy to resolve disagreements between the two initial annotators:

\begin{enumerate}
    \item \textbf{Deterministic Verification (for Facts):} Disagreements on extracted values (e.g., dates, numbers) are resolved by a third expert checking the raw text. This is treated as an objective truth problem.
    \item \textbf{Strict-Recall Principle (for Dependency):} For ambiguous reasoning cases (e.g., whether ``Call Mom'' implies a specific number), we apply the \textit{Strict-Recall Principle}. If the tool API requires a specific value (e.g., a phone number string) that is not in the query, it is marked as \textbf{\textsc{Memory-Dependent}}, even if the intent seems obvious.
    \item \textbf{Automated Leakage Check:} We utilize a rule-based filter as an auxiliary judge. If a query contains exact string matches for memory parameters, it is automatically flagged as \textbf{\textsc{Leakage}}, overriding human oversight errors.
\end{enumerate}

\section{Prompt Templates}

\newtcblisting{promptbox}[1]{
  enhanced,
  width=\linewidth,
  before=\centering,
  colback=gray!8,
  colframe=gray!60,
  boxrule=0.5pt,
  arc=2pt,
  left=6pt, right=6pt, top=6pt, bottom=6pt,
  title={#1},
  colbacktitle=gray!55,
  coltitle=white,
  fonttitle=\bfseries,
  boxed title style={sharp corners, boxrule=0pt},
  listing only,             
  listing options={         
    basicstyle=\ttfamily\small, 
    breaklines=true,        
    breakatwhitespace=false,
    columns=fullflexible,   
    keepspaces=true,        
    showstringspaces=false, 
    extendedchars=true,     
    literate={-}{-}1,       
  }
}

\begin{figure*}[t]
\begin{promptbox}{Prompt for Fact Extraction}
You are an expert information extraction system specializing in processing dialogues for fact clustering. Your primary goal is to extract key factual statements, events, and plans from the provided text.

Your output must be strictly a JSON array of objects. Do not include any other text, explanations, or formatting outside of the JSON array.

Each object in the array must contain exactly these three keys:

1. **"attribute"**: This is the **clustering key**. It must be a **specific, normalized category label** that includes the main topic and a key entity or detail. This key is used to group similar, specific facts.

   * *Example 1*: For the text "Hello, I am going to travel to Japan", the `attribute` should be **'Travel Planning (Japan)'**.
   * *Example 2*: For "I need help setting up my Azure account", the `attribute` should be **'Account Setup (Azure)'**.
2. **"facts"**: This is a concise description of the specific fact, event, or statement.

   * *Example*: For the text "Hello, I am going to travel to Japan", the `facts` description should be **'User is planning to travel to Japan'**.
3. **"source\_text"**: This is the exact dialogue snippet from the original text that supports the extracted fact.

   * *Example*: **'Hello, I am going to travel to Japan'**.

Key Instructions:

* **Attribute Normalization is Critical**: The `attribute` is the most important field. Strive for consistent, specific labels. The format should generally be `Topic (Entity)` or `Topic (Detail)`. (e.g., `Travel Planning (Japan)`, `Account Setup (Azure)`).
* **Extract Explicit Facts**: Only extract information that is explicitly stated. Do not infer or add information that is not present in the text.
* **Be Concise**: The `facts` description should be a clear, simple statement.
* **Focus on Substance**: Ignore generic conversational phrases (e.g., "hello", "thank you", "how are you") and only extract substantive facts, plans, or statements.
* **Multiple Facts**: There may be multiple facts. Please carefully extract all of them.
* **JSON Only**: The output must be a valid JSON list and nothing else.

Output example:
[
    \{\{
        "attribute": "Topic (Detail)",
        "facts": "A concise description of the fact.",
        "source_text": "The exact source text snippet."
    \}\}
]

---
Now, process the following text:
{text_content}

\end{promptbox}
\label{fig:prompt_fact_extraction}
\end{figure*}

\begin{figure*}[t]
\begin{promptbox}{Prompt for Tool Construction}
Extract parameter values from user memory to construct a tool call.

Tool Name: {tool.get('name')}
Tool Description: {tool.get('description', '')}

Complete Parameter Schema:
{json.dumps(tool_params_schema, indent=2, ensure_ascii=False)}

User Memory (extract values from here):
{json.dumps(memory_chain, indent=2, ensure_ascii=False)}

CRITICAL REQUIREMENTS:
1. At least one parameter values MUST come from the memory above (Explicit or Inferred).
2. Match parameter types exactly (string/integer/number/boolean/array/object).
3. For nested structures (arrays of objects), follow the schema's items/properties definitions.
4. All required parameters must be filled.
5. PREFER ORIGINAL TEXT: If a parameter can be found in memory, USE THE EXACT ORIGINAL TEXT from the memory.

6. PARAMETER SOURCES:
   - Memory (Explicit): Value appears verbatim in memory (e.g., "ID is 123" -> id="123").
   - Memory (Inferred): Value is inferred from context (e.g., "visiting Eiffel Tower" -> city="Paris").
   - Schema (Default): Value comes from the tool schema's default value if not found in memory.
   - Multi-hop: Parameters come from multiple different facts.

7. GROUNDING INFO:
   - For each extracted parameter, specify:
     * "source_text": The text in memory used for extraction (or "default value").
     * "type": "explicit", "inferred", or "default".

8. SOURCE IDENTIFICATION:
   - Identify which specific memory items (by their 'source_id') were necessary to derive the parameters.
   - List ALL source_ids that contributed to the parameters (including inferred ones).
   - If a parameter comes from a default value, do not include a source_id for it unless it was verified against memory.

Output JSON (follow this structure):
{json.dumps(example_output, indent=2, ensure_ascii=False)}

\end{promptbox}
\label{fig:prompt-fact}
\end{figure*}

\begin{figure*}[t]
\begin{promptbox}{Prompt for Reverse Query Generator}
You are generating a challenging, memory-dependent user query for testing an AI agent's long-term memory capabilities.

Context:
- The assistant has access to various tools and can call them to complete user requests.
- You are given a specific tool call that the assistant will make, along with the user's memory context.
- Your task is to generate a natural user query that would REQUIRE the assistant to call exactly this tool, but crucially, the user MUST OMIT key details that are already present in their memory/history.

Tool Call (the assistant will execute this):{json.dumps(tool_call, indent=2, ensure_ascii=False)}

User's Memory Context (History/Preferences):{memory_text}

Requirements for the generated query:
1. MUST BE TOOL-DEPENDENT
- The query CANNOT be answered using the model's parametric knowledge alone.
- It MUST require accessing external data or performing operations.
2. STRICTLY IMPLICIT (NO PARAMETER LEAKAGE)
- You MUST OMIT parameter values that are already established in the memory context.
- If the memory contains the destination "Dallas", the query MUST NOT say "Dallas". It should say "my destination" or "there".
- If the memory contains the date "March 13th", the query MUST NOT say "March 13th". It should say "that day" or "the date we discussed".
- The query MUST be ambiguous without the memory, but clear with the memory.
3. PROVIDE DOMAIN CONTEXT (CRITICAL)
- While avoiding specific parameter values, you MUST include enough semantic context (topic, category, or action nature) so the user knows WHICH memory thread is being referred to.
- AVOID purely generic pronouns like "it", "that", "those numbers" without any category.
- BAD: "Can you check the numbers?" (Too vague, could be anything)
- GOOD: "Can you check the atmospheric numbers?" (Better, domain is clear, but specific gas is hidden)
- GOOD: "How is the air quality data looking?" (Good, implies the topic without stating "Nitrous Oxide")
- BAD: "Book it." (Too vague)
- GOOD: "Book that flight." (Better, domain is travel)
4. REFLECT TOOL SPECIFICS (DISAMBIGUATION)
- If the tool requires a specific type of identifier (e.g., SecUID vs Username), the query should imply that specific type without stating the value.
- Example: If the tool uses 'SecUID', the query should say "using the secure ID I gave you" rather than just "my account".
- This ensures the query logically leads to the specific tool selected.
5. NATURAL AND CONVERSATIONAL
- Use phrases like "as usual", "like I said before", "for my trip", "book it", "check that thing".
- Make it sound like a continuing conversation or a user with a long history.
6. NO SYSTEM INTERNALS
- Do NOT mention "tool", "function", "API", "call", "parameter".

Output Format:
Return ONLY a valid JSON object with one field:
{{
"query": "the generated user query here"
}}

Do not include any explanations, comments, or text outside this JSON object.
        
\end{promptbox}
\label{fig:promp_reverse_query_generator}
\end{figure*}

\end{document}